\documentclass{article}

\usepackage[numbers, compress]{natbib}

\usepackage[preprint]{neurips_2025}

\usepackage[utf8]{inputenc} 
\usepackage[T1]{fontenc}    
\usepackage{hyperref}       
\usepackage{url}            
\usepackage{booktabs}       
\usepackage{amsfonts}       
\usepackage{nicefrac}       
\usepackage{microtype}      
\usepackage{xcolor}         
\usepackage[ruled,vlined]{algorithm2e}
\usepackage{graphicx}
\usepackage{caption}
\usepackage{threeparttable}
\usepackage{multirow}
\usepackage{amsmath} 
\usepackage{enumitem}
\usepackage{wrapfig}
\usepackage{pifont}
\usepackage{makecell}

\title{Know³-RAG: A Knowledge-aware RAG Framework with Adaptive Retrieval, Generation, and Filtering}

\author{%
  Xukai Liu$^{1,2}$, Ye Liu$^{1,2}$, Shiwen Wu$^{3}$, Yanghai Zhang$^{1,2}$, \\
  \textbf{Yihao Yuan$^{1}$, Kai Zhang$^{1,2}$, Qi Liu$^{1,2}$} \\
  $^{1}$School of Computer Science and Technology, University of Science and Technology of China\\
  $^{2}$State Key Laboratory of Cognitive Intelligence, Hefei, Anhui, China \\
  $^{3}$The Hong Kong University of Science and Technology\\
  \{chthollylxk, yhzhang0612, yh0319\}@mail.ustc.edu.cn, yeliu.liuyeah@gmail.com\\
  swubs@connect.ust.hk, \{kkzhang08, qiliuql\}@ustc.edu.cn, \\
}

\begin{document}

\maketitle

\begin{abstract}
  Recent advances in large language models (LLMs) have led to impressive progress in natural language generation, yet their tendency to produce hallucinated or unsubstantiated content remains a critical concern. To improve factual reliability, Retrieval-Augmented Generation (RAG) integrates external knowledge during inference. However, existing RAG systems face two major limitations: (1) unreliable adaptive control due to limited external knowledge supervision, and (2) hallucinations caused by inaccurate or irrelevant references. To address these issues, we propose \textbf{Know³-RAG}, a knowledge-aware RAG framework that leverages structured knowledge from knowledge graphs (KGs) to guide three core stages of the RAG process, including retrieval, generation, and filtering. Specifically, we introduce a knowledge-aware adaptive retrieval module that employs KG embedding to assess the confidence of the generated answer and determine retrieval necessity, a knowledge-enhanced reference generation strategy that enriches queries with KG-derived entities to improve generated reference relevance, and a knowledge-driven reference filtering mechanism that ensures semantic alignment and factual accuracy of references. Experiments on multiple open-domain QA benchmarks demonstrate that Know³-RAG consistently outperforms strong baselines, significantly reducing hallucinations and enhancing answer reliability.
\end{abstract}

\section{Introduction}
\label{intro}
Large language models (LLMs) have demonstrated remarkable performance across a wide range of natural language processing (NLP) tasks, such as machine translation, text generation, and question answering~\cite{raiaan2024LLM_application_survey}. However, they might generate content that is factually incorrect or unsubstantiated — a phenomenon known as hallucination. As LLMs become increasingly integrated into high-stakes domains, reducing hallucinations has emerged as a central challenge for building reliable language generation systems~\cite{ji2023survey_hallucination}.

\begin{figure}[t]
    \centering
    \vspace{-0.7cm}
    \includegraphics[width=\textwidth]{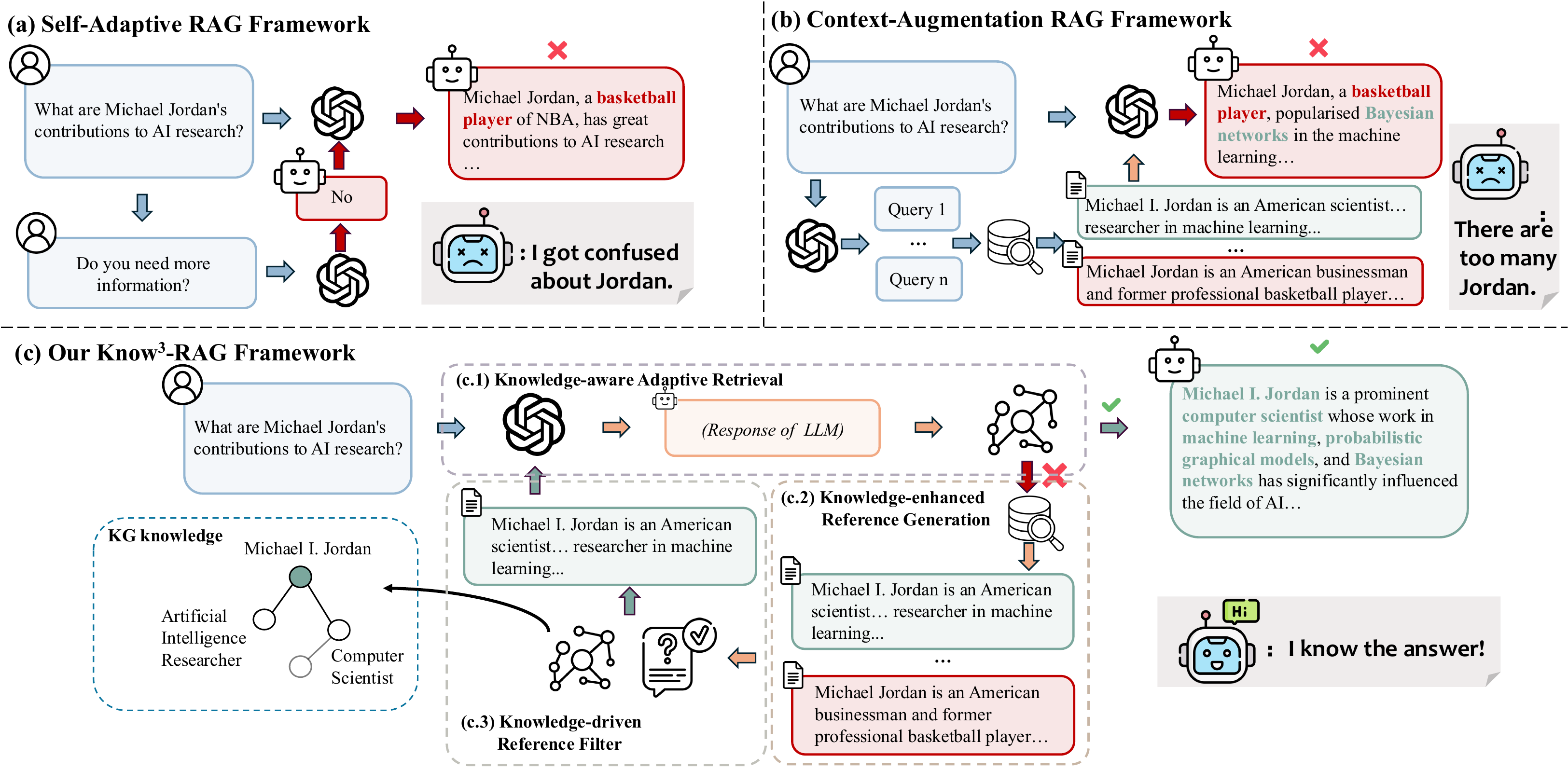}
    \caption{Comparison of three different RAG frameworks. Our proposed Know³-RAG Framework employs knowledge graphs to facilitate adaptive retrieval and reference processing.}
    \label{fig:case}
    \vspace{-0.3cm}
\end{figure}

To address this issue, Retrieval-Augmented Generation (RAG), a paradigm that integrates external knowledge during inference, has emerged as a popular method to mitigate hallucinations~\cite{RAG2024survey}. Recent studies~\cite{feng2024knowledge-card,li-etal-2024-retrieval-or-lc, yu2023GenRead} have explored various forms of LLM-generated content as external knowledge to further enhance the performance of RAG. Specifically, these efforts can be broadly categorized into two directions. 
The first (Figure~\ref{fig:case}(a)) explores self-adaptive RAG, where the model learns to determine whether external evidence is needed. Typical approaches involve model self-ask~\cite{wang2023SKR, asai2023self-rag} or training predictive modules to estimate retrieval necessity~\cite{jeong2024Adaptive-RAG, zhao2023thrust}.
The second (Figure~\ref{fig:case}(b)) investigates context-augmentation RAG, which aims to improve the quality of the information acquired via LLMs. 
This is achieved by enhancing query expressiveness through LLMs~\cite{wang2023query2doc, wang2024blendfilter} or synthesizing high-quality reference content directly from the LLMs' internal knowledge~\cite{sunrecitation, feng2024knowledge-card}.

Despite their effectiveness, existing methods still encounter two major limitations. Firstly, for the self-adaptive RAG, their adaptive retrieval mechanisms are typically guided by internal model signals without knowledge verification, making them susceptible to training data biases and thus limiting their generalization ability. In fact, studies have shown that self-ask signals often fail to accurately reflect true retrieval needs~\cite{zhang2024retrievalqa}. An example can be seen in Figure~\ref{fig:case}(a), where the self-ask part refuse to explore more information about \textit{Michael Jordan}, resulting in its confusion in the answer.

Secondly, for context-augmentation RAG, current pipelines tend to emphasize relevance during retrieval while overlooking the quality control, leading to the inclusion of noisy or misleading content. On the one hand, retrievers may select factually incorrect content due to a preference for fluent LLM-generated passages even if they are incorrect~\cite{IR_and_LLM_bias}. On the other hand, topically irrelevant information may also be retrieved, thereby confusing the LLM's understanding of the context. As shown in Figure~\ref{fig:case}(b), the basketball player \textit{Michael Jordan} is incorrectly included in the retrieval results of computer scientists, thereby misleading the model generation process. Such failures highlight the need for external knowledge supervision throughout the RAG pipeline, from adaptive control to reference filtering.

To address these limitations, we propose \textbf{Know³-RAG}, a Knowledge-aware RAG framework that leverages structured knowledge graphs (KGs) as external supervision across three core stages of the RAG process, including retrieval, generation and filtering. Specifically, as shown in Figure~\ref{fig:case} (c), to address the problem of unreliable adaptive retrieval, we first proposed the Knowledge-aware Adaptive Retrieval, which employs KG representation to assess the confidence of current generated answer and determine whether additional retrieval is necessary. Subsequently, to boost the quality of references, we introduce Knowledge-enhanced Reference Generation, which injects KG-related entities into the query formulation, thus generating more relevant and specific reference documents. Further, a Knowledge-driven Reference Filtering mechanism is designed to evaluate the retrieved references for both semantic relevance and factual consistency with the given input query. 

Finally, extensive experiments on multiple open-domain QA benchmarks demonstrate that Know³-RAG achieves state-of-the-art performance while significantly improving interpretability and reducing hallucinations. The code is available at \url{https://github.com/laquabe/Know3RAG}.


\section{Related Works}
\subsection{Retrieval-Augmented Generation}
Retrieval-Augmented Generation (RAG) has emerged as a key strategy for mitigating hallucinations in large language models (LLMs) by integrating external knowledge into inference~\cite{RAG2024survey}.
Initial methods, such as In-Context RALM~\cite{ram2023In-Context-RALM}, directly incorporated retrieved text into the prompts, while IRCoT~\cite{trivedi2023IRCoT} integrated iterative retrieval with reasoning steps.
Previous RAG methods often retrieve a fixed number of documents, risking either inadequate coverage (if too few are retrieved) or noise introduction (if too many are retrieved), both of which can harm model effectiveness~\cite{huang2025hallucination_survey}.
To address this, adaptive retrieval techniques were developed. For instance, SKR~\cite{wang2023SKR} introduced a self-ask mechanism to guide retrieval. Adaptive RAG~\cite{jeong2024Adaptive-RAG} and Thrust~\cite{zhao2023thrust} leveraged a classifier to determine retrieval necessity, which was based on text or embedding,  respectively. Self-RAG~\cite{asai2023self-rag} trained the LLM to emit special tokens indicating retrieval actions. However, their reliance on the LLM itself or the training data makes their generalization susceptible to data biases.

Meanwhile, research also focused on optimizing the content provided to the LLMs. Some methods aimed to retrieve more relevant information by enhancing query expressiveness. Query2Doc~\cite{wang2023query2doc} and HyDE~\cite{gao2023HyDE} expanded the query by rewriting LLMs. ToC~\cite{kim2023ToC} employs a tree-structured search to explore diverse evidence. BlendFilter~\cite{wang2024blendfilter} combined internal and external query augmentations with a LLM-based knowledge filtering mechanism to improve retrieval quality.
Other works concentrate on generating high-quality references directly from LLMs' internal knowledge. Recitation~\cite{sunrecitation} and GenRead~\cite{yu2023GenRead} utilize LLMs to synthesize pseudo-documents as reference documents. LongLLMLingua~\cite{jiang-etal-2024-longllmlingua} compresses documents via a trainable model to extract key information. DSP~\cite{li2023DSP} and BGM~\cite{ke2024BGM} further optimize the model through reinforcement learning based on LLM feedback. Knowledge Card~\cite{feng2024knowledge-card} suggested fine-tuning domain knowledge models and allowing the LLM to select the useful domain. 
Nevertheless, these approaches often do not integrate explicit verification mechanisms to ensure the quality of external knowledge.
In contrast, by providing explicit verification powered by knowledge graphs, our proposed training-free method precisely addresses this gap, significantly enhancing the relevance and reliability of the references.

\subsection{LLM Reasoning with Knowledge Graph}
Knowledge Graphs (KGs), like DBpedia~\cite{lehmann2015dbpedia}, YAGO~\cite{rebele2016yago}, and Wikidata~\cite{vrandevcic2014wikidata}, explicitly structure rich factual knowledge. Integrating this structured knowledge provides a promising approach to enhance LLM reasoning capabilities~\cite{pan2024KGLLMsurvey}. Some methods focus on adapting KG information for LLM reasoning. For example, Mindmap~\cite{wen2024mindmap} converted KGs into mind maps via designed prompts to facilitate model comprehension. StructGPT~\cite{jiang2023structgpt} and React~\cite{yao2023react} leveraged the Tool-augmented LLM paradigm, defining APIs to enable LLM interaction with KGs. Furthermore, Think-on-Graph~\cite{think-on-graph} regarded the LLM as an agent exploring KGs, while KnowledgeNavigator~\cite{guo2024knowledgenavigator} identifies similar questions to provide diverse starting points.

Other methods focus on integrating internal model knowledge with external KG evidence. For instance, CoK~\cite{wang-etal-2024CoK} assumed that LLMs have internalized the KG knowledge KG and prompts it to generate relevant triples before answering the question. RoG~\cite{RoG} and Readi~\cite{cheng2024Readi} generated relational paths via fine-tuning or prompting, and leverage these paths to reason over KGs.
R3~\cite{toroghi2024R3} utilized KGs to validate individual reasoning steps performed by the LLM. GraphRAG~\cite{edge2024graphrag} employed LLMs to construct KGs from unstructured documents and leverages the KGs to enhance retrieval.
However, many existing methods primarily engage with local, textual information from the KG, overlooking the rich global structural information inherent in the graph. Our framework, in contrast, addresses this limitation by employing Knowledge Graph Embedding (KGE) techniques. This allows us to encode global structural and semantic information from the KG and integrate it throughout the retrieval and generation pipeline, thereby improving the relevance and consistency of the generated references.

\section{Problem Definition}
We consider the task of open-domain question answering, which aims to generate an answer to a given question without providing any predefined context in advance. Formally, given a question \( Q\), the model seeks to produce an answer \( a \) by leveraging any available external knowledge sources.

Following the setting of existing paradigm~\cite{sunrecitation,yu2023GenRead,feng2024knowledge-card}, we adopt the LLM-generated content as the external knowledge, which is motivated by the observation that LLM outperforms traditional retrieval methods in handling complex queries~\cite{li-etal-2024-retrieval-or-lc}.
Speically, we decomposes the task into two stages: (1) generating a set of references \( \mathcal{D} = \{d_1, d_2, \ldots, d_k\} \) relevant to \( Q \), and (2) generating the answer \( a \) based on the question and the references. The incorporation of external knowledge helps mitigate the limitations of parametric memory in models, enabling more accurate answer generation.

\section{Know³-RAG Method}
\label{method}

\subsection{Overview}

\begin{figure}[t]
    \centering
    \vspace{-1.2cm}
    \includegraphics[width=\textwidth]{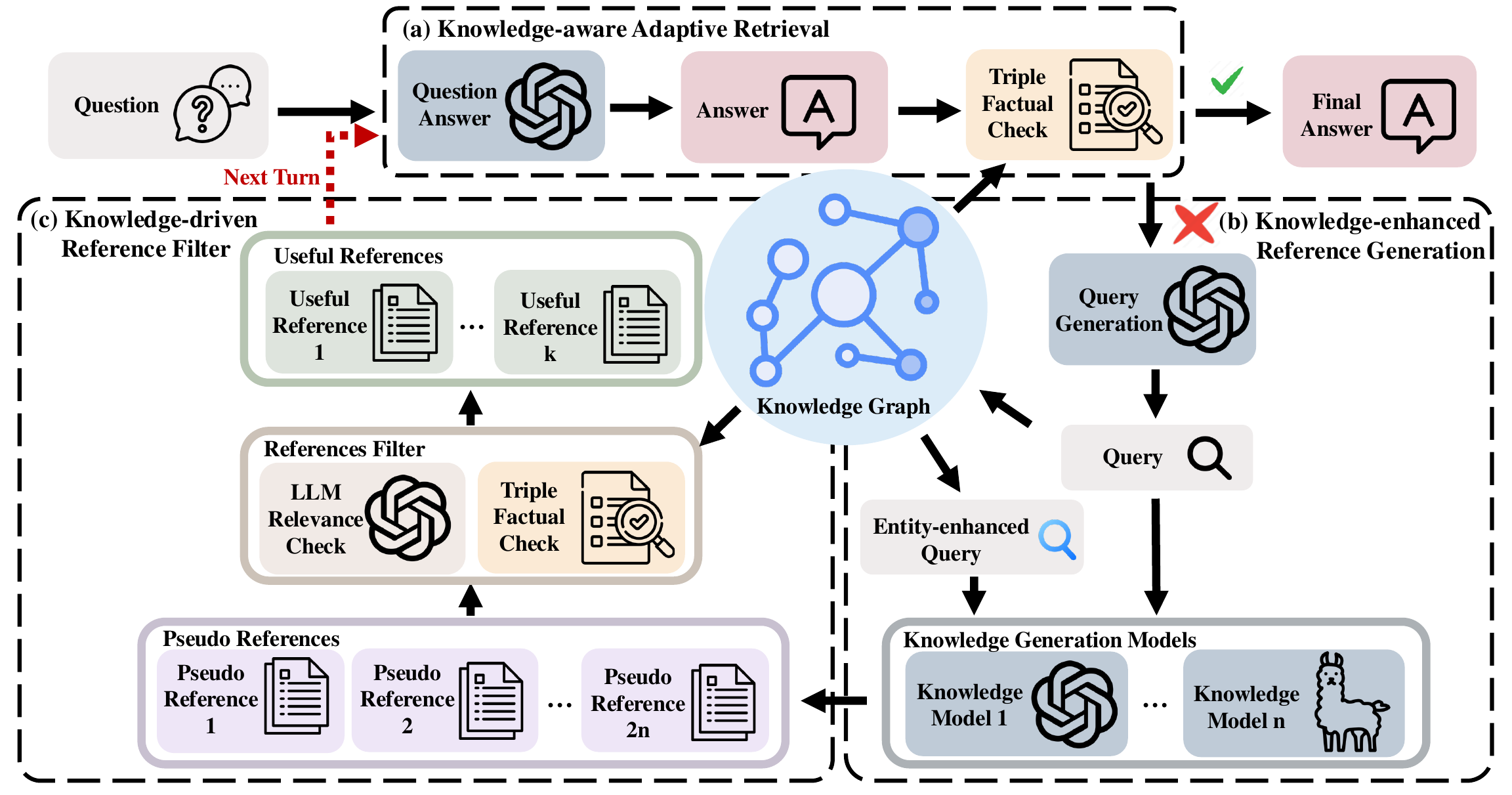}
    \caption{Model Architecture of Know³-RAG, which contains: (a) Knowledge-aware Adaptive Retrieval, (b) Knowledge-enhanced Reference Generation, (c) Knowledge-driven Reference Filter.}
    \label{fig:framework}
\end{figure}

As illustrated in Figure~\ref{fig:framework}, our Know³-RAG framework comprises three core components: (a) Knowledge-aware Adaptive Retrieval, (b) Knowledge-enhanced Reference Generation, and (c) Knowledge-driven Reference Filter. These modules are organized in a closed-loop architecture to enable iterative refinement of answers.
In detail, the Knowledge-aware Adaptive Retrieval (a) receives the answer from the previous QA iteration and verifies the reliability of extracted triples in the answer using a knowledge graph (KG), thereby determining the necessity for additional retrieval. 
If further retrieval is required, the Knowledge-enhanced Reference Generation (b) augments the query with relevant KG entities and then generates pseudo-references via knowledge generation models. 
Subsequently, the Knowledge-driven Reference Filter (c) evaluates the pseudo-references for relevance and factuality and retains the useful references among them. 
The filtered references are then combined with the original question and previous references to inform the QA model to generate the next round answer, thus completing one full process cycle.

\subsection{Knowledge-aware Adaptive Retrieval}
\label{method:adaptive}

As discussed in the Introduction, the adaptive iterative mechanism determines whether to invoke retrieval based on the reliability of the current answer. The central challenge lies in evaluating the reliability of the generated answer without human annotations.
To address this, we employ the Knowledge Graph Embedding (KGE) model to quantify the reliability of the generated text. Specifically, for the answer \( a_t \) generated at the \( t \)-th iteration for the question \( Q \), we first extract a set of evidence triples \( \text{Tri}^{(a_t)} = \{\text{tri}_1^{(a_t)}, \ldots, \text{tri}_i^{(a_t)}\} \) using a large language model. Each triple is then scored using a relative triple scoring function, and the overall reliability score \( s_t \) for the answer is computed by aggregating the individual triple scores. Formally:
\begin{equation}
    a_t = \text{LLM}_{\text{ans}}(Q, D_t^c), \ \ \  \text{Tri}^{(a_t)} = \text{LLM}_{\text{tri}}(a_t),\ \ \ s_t = \sum_i S(\text{tri}_i^{(a_t)}),
\end{equation}
where \( \text{LLM}_{\text{ans}} \) denotes the QA model, and \( D_t^c \) is the set of references available at iteration \( t \). \( \text{LLM}_{\text{tri}} \) refers to the triple extraction LLM, and \( S(\cdot) \) is the relative triple scoring function.

\textbf{Relative Triple Scoring.} Scores produced by KGE models lack absolute interpretability and are meaningful only in a comparative sense, as they are primarily designed for ranking tasks~\cite{bordes2013TransE,trouillon2016complex}.
To leverage these scores for reliability assessment, we propose a relative triple score.
Specifically, for each extracted triple \( \text{tri}_i^{(a_t)} \) from the generated answer \( a_t \), we retrieve a set of reference triples \( \text{Tri}_{\text{ref}_i}^{(a_t)} = \{\text{tri}_{\text{ref}_{i1}}^{(a_t)}, \ldots, \text{tri}_{\text{ref}_{ij}}^{(a_t)}\} \) from the KG that share the same head entity \( h_i^{(a_t)} \). We then compute the triple relative score by measuring the absolute value between its KGE score and the average KGE score of its reference triples:
\begin{equation}
S(\text{tri}_i^{(a_t)}) = \left| \text{KGE}(\text{tri}_i^{(a_t)}) - \frac{1}{|\text{Tri}_{\text{ref}_i}^{(a_t)}|} \sum_j \text{KGE}(\text{tri}_{\text{ref}_{ij}}^{(a_t)}) \right|,
\end{equation}
where \( \text{KGE}(\cdot) \) denotes the KGE scoring function, and \( |\cdot| \) is the absolute value function. Intuitively, a lower relative score indicates higher consistency between the generated content and knowledge in the KG, thus reflecting greater reliability of the answer.
During inference, our adaptive retrieval employs an iterative process with a dynamically increasing threshold \( \theta_t \)~\cite{xu2021dash, sun-etal-2019-pullnet}. At each iteration \( t \), we compare the relative triple score $s_t$ against $\theta_t$ to decide whether to terminate. The threshold is defined as \( \theta_t = \theta_0 \left(\frac{c}{1 + e^{1-\theta_0}}\right)^t \), where \( \theta_0 \) is a hyperparameter dependent on the QA model and dataset, and \( c \) is a constant. To prevent infinite iteration, a maximum number of iterations, denoted by $\mathcal{T}$, is set. If \( s_t < \theta_t \) or if $t = \mathcal{T}$, the answer is output; otherwise, a reference query is triggered for further refinement.

\subsection{Knowledge-enhanced Reference Generation}
\label{method:reference generation}
When retrieval is triggered, the Knowledge-enhanced Reference Generation introduces additional references to support answer generation. Specifically, we first generate a new query \( q_t \) for the current iteration, defined as:
\begin{equation}
q_t = 
\begin{cases}
Q, & t = 1 \\
\text{LLM}_{\text{query}}(Q, D_t^c, a_t), & t > 1,
\end{cases}
\end{equation}
where \( \text{LLM}_{\text{query}} \) denotes a LLM for query generation. After obtaining the query \( q_t \), we further enhance it by incorporating entity information from the knowledge graph (KG) to enrich the query. Specifically, we get a set of relevant entities \( E_t \), including three types:
\begin{itemize}[leftmargin=*]
\item  \textbf{Query linked entities \( E_t^q \)}. Entities directly linked to the query via entity linking, representing knowledge most relevant to the query.
\item \textbf{Local neighbor entities \( E_t^l \)}. Entities that are one-hop neighbors of \( E_t^q \) in the KG. We filter these neighbors based on the semantic relevance between their relations and the query.
\item  \textbf{Global predicted entities \( E_t^g \)}. Tail entities predicted by the KGE model along relations connected to \( E_t^q \), supplementing \( E_t^l \) to mitigate knowledge graph incompleteness.
\end{itemize}
Incorporating these relevant entities enables more effective knowledge infusion, improving the efficiency of the iterative generation process. The detailed entity enhancement procedure is presented in Algorithm~\ref{alg:algorithm1} (refer to Appendix~\ref{appendix:algorithm}). The entity-enhanced query \( q_t^{\text{KG}} \) is then constructed by concatenating the relevant entity set and the original query: $q_t^{\text{KG}} = [E_t; q_t]$. Then, using both \( q_t \) and \( q_t^{\text{KG}} \), we generate candidate reference documents through knowledge models, which are any kind of generative models. Unlike traditional retrieval methods, knowledge models can generate more targeted documents for complex queries~\cite{sunrecitation, yu2023GenRead, li-etal-2024-retrieval-or-lc}. The document generation process is formalized as:
\begin{equation}
d = K(q),
\end{equation}
where \( K(\cdot) \) denotes a knowledge model. To further enrich the references, we adopt a multi-source strategy, invoking \( n \) different knowledge models to generate documents based on both \( q_t \) and \( q_t^{\text{KG}} \), resulting in an initial reference candidate set \( D_m \):
\begin{equation}
D_m = \{ d_{t_1}, \ldots, d_{t_n}, d_{t_1}^{\text{KG}}, \ldots, d_{t_n}^{\text{KG}} \},
\end{equation}
\subsection{Knowledge-driven Reference Filter}

As discussed in Introduction, noisy or misleading content may be included to influence the generation process.
To mitigate the risk of introducing irrelevant or inaccurate documents into the final reference set, we design the Knowledge-driven Reference Filtering module, which consists of two core components: (1) LLM relevance check and (2) triple factual check.

Specifically, the LLM relevance check assesses the semantic relevance between the generated document \( d \) and the original question \( Q \). To improve the robustness of the relevance assessment, we incorporate entity information \( E^Q \) extracted from the question as additional context. Formally, the relevance check is defined as:
\begin{equation}
c = \text{LLM}_{\text{rel}}(d, Q, E^Q),
\end{equation}
where \( \text{LLM}_{\text{rel}} \) denotes a LLM used for relevance judgment, and \( c \in \{\text{False}, \text{True}\} \) represents the binary relevance decision.
The triple factual check follows the approach introduced in the Knowledge-aware Adaptive Retrieval module: we extract triples from the document and score the reliability of the document using the relative triple score.

Finally, we integrate the results from both modules to filter useful references. Specifically, among documents judged as relevant, we select the top \( k \) documents ranked by their relative triple scores, forming the next iteration’s reference set \( D_{t+1}^c \):
\begin{equation}
\begin{aligned}
    D_{t+1}^c &= D_t^c \cup \{ d_{t_1}^c, \ldots, d_{t_k}^c \},\\
    a_{t+1} &= \text{LLM}_{\text{ans}}(Q, D_{t+1}^c),
\end{aligned}
\end{equation}
where \( d_{t_k}^c \) denotes the documents selected in the current round. In this way, our Know³-RAG framework completes a full circle.

\section{Experiments}
\label{exp}
\subsection{Datasets}
\label{exp:datasets}
\begin{wraptable}{r}{0.4\linewidth}
\centering
\vspace{-1em}
\caption{The Statistics of Test Datasets}
\label{table:dataset}
\begin{tabular}{lc}
\toprule
\textbf{Dateset} & \textbf{Nums.}  \\ \midrule
HotpotQA & 7,405 \\
2WikiMultiHopQA & 12,576 \\
PopQA & 14,267 \\
\bottomrule
\end{tabular}
\end{wraptable}
We evaluate our approach on three widely used open-domain QA benchmarks: HotpotQA~\cite{yang2018hotpotqa}, 2WikiMultiHopQA~\cite{ho-etal-20202WikiMultiHopQA}, and PopQA~\cite{yang2018hotpotqa}. Following the setup in Wang et al.~\cite{wang2024blendfilter}, we report results on the publicly available development sets for HotpotQA and 2WikiMultiHopQA, and on the official test set for PopQA. We use Exact Match (EM) and F1 as evaluation metrics. Dataset statistics are summarized in Table~\ref{table:dataset}.
It is important to note that although these datasets may provide auxiliary information (e.g., supporting passages), we do not use any of it during evaluation. All methods rely solely on the input question, and all additional context is obtained by the method itself.

\subsection{Baselines}
\label{exp:baseline}
We adopt following state-of-the-art baselines to evaluate our proposed method:
\begin{itemize}[leftmargin=*]
\item \textbf{Direct Prompting}~\cite{brown2020direct_prompt}: A simple approach where the LLM directly answers questions without reasoning steps. We evaluate both directly answering with and without retrieval.
\item \textbf{Chain-of-Thought}~\cite{wei2022CoT}: This method instructs the LLM to generate reasoning steps before answering. Similar to Direct Prompting, we evaluate both CoT with and without retrieval.
\item \textbf{Recite}~\cite{sunrecitation}: This method leverages the LLM’s parametric knowledge by prompting it to first recite relevant factual content, then LLMs respond based on that recitation.
\item \textbf{SKR}~\cite{wang2023SKR}: A self-adaptive RAG that queries the LLM to decide whether retrieval is necessary.
\item \textbf{Knowledge Card (KC)}~\cite{feng2024knowledge-card}: This approach uses domain-finetuned knowledge models for document generation. We evaluate the three KG-related knowledge models provided by the authors—Wikidata, Wikipedia, and YAGO—as external knowledge sources.
\item \textbf{Chain of Knowledge (CoK)}~\cite{wang-etal-2024CoK}: Based on the hypothesis that LLMs internalize knowledge graph, this method prompts the LLM to output relevant triples before generating an answer.
\item \textbf{BlenderFilter (BF)}~\cite{wang2024blendfilter}: A hybrid retrieval framework that combines internal knowledge of LLMs and external knowledge of retrieval. Retrieved content is filtered by an LLM to discard irrelevant information, aiming to broaden retrieval coverage while maintaining quality.
\end{itemize}
Following Wang et al.~\cite{wang2024blendfilter}, for all retrieval-based methods, we use ColBERTv2~\cite{santhanam2022colbertv2} as the retriever and employ the 2017 Wikipedia abstract dump as the retrieval corpus~\cite{yang2018hotpotqa}.
We exclude the KBQA methods such as Think-on-Graph~\cite{think-on-graph}, as they need to access the gold topic entities and structured queries, which are incompatible with our open-domain QA setting. 

\subsection{Implementation Details}
\label{exp:imp details}

We evaluate our framework using three types of LLMs: GLM4-9B~\cite{glm2024chatglm}, Qwen2.5-32B~\cite{qwen2.5}, and GPT-4o-mini, representing two open-source models of different sizes and one closed-source model. For entity linking, we use spaCy to efficiently identify entities. We choose ComplEX~\cite{trouillon2016complex} as the KGE model.
For the knowledge models, we consider two categories:  
(1) Non-fine-tuned models, including LLaMA3-8B~\cite{grattafiori2024llama}, Qwen2.5-7B~\cite{qwen2.5}, and the QA model itself;  
(2) Fine-tuned domain models from Knowledge Card~\cite{feng2024knowledge-card}, based on OPT-1.3B~\cite{zhang2022opt} and trained separately on Wikidata, Wikipedia, and YAGO. 
To maintain efficiency, each knowledge model outputs only one document per query.
All LLM-based modules apart from the QA model use lightweight open-source LLMs like LLaMA3-8B. The hyperparameter \( \theta_0 \) is in Table~\ref{table:theta} in Appendix~\ref{app:theta}, the constant c is 128, and the maximum number of iteration $\mathcal{T}$ is 2. At each iteration, we retain \( k + t \) documents as useful references, with \( k \) set to 5.
All QA models are prompted using a 3-shot in-context learning setup and set the temperature to 0. Detailed prompts are provided in Appendix~\ref{Appendix:prompt}.

\subsection{Experimental Results}
Table~\ref{tab:main} presents the main results across three models and three open-domain QA datasets. Our Know³-RAG framework consistently achieves the best or near-best performance in both EM and F1 scores, demonstrating robust generalizability across models and datasets. 

We also highlight two interesting observations. First, both intrinsic and external knowledge are essential. For example, Chain-of-Knowledge (CoK), which relies on intrinsic knowledge, outperforms BlenderFilter (BF), which leverages external knowledge, on the 2WikiMultiHopQA (2Wiki) dataset, while the opposite holds for PopQA. In contrast, our method integrates both to achieve superior performance across most models and datasets. Second, we find that naive self-ask strategies, such as SKR, may misjudge the actual retrieval need, leading to an inferior result compared to direct retrieval (RAG) in some scenarios. Our Know³-RAG framework addresses this issue through our triple-based adaptive RAG, which more accurately identifies the knowledge gaps of the model and iteratively refines the quality of the answers.

\begin{table}[t] 
  \centering
  \caption{Main Results} 
  \label{tab:main} 
  \setlength{\tabcolsep}{3pt} 
    {\small
    \resizebox{\textwidth}{!}{
  \begin{tabular}{@{}l |cc| cc| cc| cc| cc| cc| cc| cc| cc@{}} 
    \toprule
    & \multicolumn{6}{c}{\textbf{GLM4-9b}} & \multicolumn{6}{c}{\textbf{Qwen2.5-32b}} & \multicolumn{6}{c}{\textbf{GPT4o-mini}} \\
    \cmidrule(lr){2-7} \cmidrule(lr){8-13} \cmidrule(lr){14-19} 
     \multicolumn{1}{@{}l|}{\textbf{Method}}  & \multicolumn{2}{c}{\textbf{HotPot}} & \multicolumn{2}{c}{\textbf{2Wiki}} & \multicolumn{2}{c}{\textbf{Pop}} & \multicolumn{2}{c}{\textbf{HotPot}} & \multicolumn{2}{c}{\textbf{2Wiki}} & \multicolumn{2}{c}{\textbf{Pop}} & \multicolumn{2}{c}{\textbf{HotPot}} & \multicolumn{2}{c}{\textbf{2Wiki}} & \multicolumn{2}{c}{\textbf{Pop}} \\
    \cmidrule(lr){2-3} \cmidrule(lr){4-5} \cmidrule(lr){6-7} \cmidrule(lr){8-9} \cmidrule(lr){10-11} \cmidrule(lr){12-13} \cmidrule(lr){14-15} \cmidrule(lr){16-17} \cmidrule(lr){18-19} 
     & EM & F1 & EM & F1 & EM & F1 & EM & F1 & EM & F1 & EM & F1 & EM & F1 & EM & F1 & EM & F1 \\
    \midrule
    Direct       & 0.181 & 0.242 &  0.252 & 0.278 & 0.185 & 0.222 & 0.237 & 0.322 & 0.277 & 0.322 & 0.181 & 0.237 & 0.220 & 0.294 & 0.226 & 0.272 & 0.327 & 0.378 \\
    CoT          & 0.190 & 0.262 &  0.245 & 0.308 & 0.185 & 0.235 & 0.233 & 0.331 & 0.243 & 0.317 & 0.190 & 0.237 & 0.315 & 0.426 & \textbf{0.383} & 0.443 & 0.355 & 0.401 \\
    RAG-D   &  0.219 &  0.281 & 0.233  & 0.300  & 0.279  & 0.326  & 0.248  & 0.335  & 0.269  & 0.324  & 0.299  &  0.352 & 0.291  & 0.383  & 0.301  & 0.360  & 0.316  & 0.362  \\
    RAG-C  &  0.196 & 0.262  & 0.193  &  0.242 & 0.257  & 0.302  & 0.243  & 0.321  &  0.262 & 0.304  & 0.273  & 0.327  &  0.307 &  0.404 &  0.337 & 0.410  & 0.373  & 0.428 \\
    Recite   & 0.215 & 0.318 & 0.209 & 0.274 & 0.183 & 0.240 &  0.243 & 0.345 & 0.273 & 0.339 & 0.193 & 0.248 & 0.328 & 0.447 & 0.347 & 0.434 & 0.331 & 0.386 \\
    SKR  & 0.218  &  0.280 & 0.260  & 0.312  & 0.286 & 0.332  &  0.253 & 0.336  &  0.232 & 0.280  & 0.232  & 0.282  & 0.303  &  0.402 & 0.232  &  0.280 & 0.366  & 0.420  \\
    KC & 0.183 & 0.262 & 0.229  & 0.261  & 0.171  & 0.208  & 0.221  &  0.295 &  0.167 &  0.195 &  0.279 & 0.309  & 0.313 &  0.410 & 0.172  & 0.208  &  0.297 &   0.331  \\
    CoK &  0.205 & 0.288 & \textbf{0.269}  & 0.302  & 0.192  & 0.230  & 0.251  & 0.353  & 0.274  & 0.333  & 0.200  & 0.245  &  0.328 & 0.443  & 0.379  & 0.441  &  0.337 &  0.387 \\
    BF &  0.221 &  0.308 & 0.232  & 0.273 &  0.287 &  0.332 &  0.257 &  0.343 & 0.207  & 0.252  &  0.296 & 0.349  & \textbf{0.339}  & 0.448  &  0.272 &  0.339 & 0.380  & 0.439  \\
    \midrule 
    \textbf{Know³-RAG} & \textbf{0.235} & \textbf{0.351} & 0.261  & \textbf{0.362}  & \textbf{0.337}  & \textbf{0.399}   & \textbf{0.270}   &  \textbf{0.387}  &  \textbf{0.343}  & \textbf{0.424}  & \textbf{0.332}  & \textbf{0.406}  & 0.335  & \textbf{0.453}  & 0.382  & \textbf{0.457}  & \textbf{0.396}  & \textbf{0.462}  \\
    \bottomrule
  \end{tabular}
  }
  }
\end{table}

\subsection{Ablation Study}

\begin{wraptable}{r}{0.555\linewidth}
\centering
\vspace{-4.5em}
  \caption{Ablation Analysis on GLM4-9b} 
  \label{tab:ablation} 
  \setlength{\tabcolsep}{3pt} 
    {\small
  \begin{tabular}{@{}l |cc| cc| cc@{}} 
    \toprule
    \multirow{2}{*}[-0.7ex]{\textbf{Method}} & \multicolumn{2}{c}{\textbf{HotPot}} & \multicolumn{2}{c}{\textbf{2Wiki}}  & \multicolumn{2}{c}{\textbf{Pop}} \\
    \cmidrule(lr){2-3} \cmidrule(lr){4-5}  \cmidrule(lr){6-7} 
     & EM & F1 & EM & F1 & EM & F1 \\
    \midrule
    Know³-RAG & 0.235 & 0.351 &  0.261 & 0.362 &  0.337 & 0.399 \\
    \midrule
    \ding{192} w/o Reference & 0.190 & 0.262 &  0.245 &  0.308 &  0.185 & 0.235 \\
    \quad w/o KG query & 0.212 & 0.321 &  0.253 & 0.341 & 0.228 & 0.284\\
    \quad w/o raw query & 0.193 & 0.287 & 0.207 & 0.267 & 0.328 & 0.385 \\
    \midrule
    \ding{193} w/o Filter & 0.221 & 0.335 & 0.229 & 0.322 & 0.314 &  0.380 \\
    \quad w/o Tri-check & 0.230 & 0.347 & 0.231 & 0.324 & 0.331 & 0.395 \\
    \quad w/o Rel-check & 0.211 & 0.318 & 0.228 & 0.315 & 0.296 & 0.359\\
    \midrule
    \ding{194} w/o  \multirow{2}{*}{\makecell[l]{Adaptive \\ Retrieval}} 
    & \multirow{2}{*}{0.227} & \multirow{2}{*}{0.339} & \multirow{2}{*}{0.259} &  \multirow{2}{*}{0.357} & \multirow{2}{*}{0.330} & \multirow{2}{*}{0.391} \\ 
     & & & & & \\ 

    \bottomrule
  \end{tabular}
  }
\vspace{-2em}
\end{wraptable}

Table~\ref{tab:ablation} reports the results of the ablation study conducted on GLM4-9b across three QA datasets. The results confirm the effectiveness of all three proposed modules\textsuperscript{\ding{192}\ding{193}\ding{194}}. We further provide a detailed analysis of two components.
For the Knowledge-enhanced Reference Generation module, intrinsic knowledge proves more valuable than external knowledge. This observation suggests that large language models increasingly internalize factual knowledge as they develop. Moreover, it implies that incorporating knowledge graphs may introduce noise, such as linking to the wrong entities, which can misguide reference generation.
As for the Knowledge-driven Reference Filter module, we find that relevance check yields greater improvements than factual check. This indicates that factually correct yet contextually irrelevant content poses a greater risk of misleading the model’s reasoning, highlighting the importance of semantic alignment in the reference selection process.

\subsection{Comparison of Knowledge Models}

\begin{wraptable}{r}{0.575\linewidth}
\centering
\vspace{-3em}
  \caption{Performance of Partial Knowledge Models} 
  \label{tab:knowledge base} 
  \setlength{\tabcolsep}{3.5pt} 
    {\small
  \begin{tabular}{@{}l |cc| cc| cc@{}} 
    \toprule
     \multirow{2}{*}[-0.7ex]{\textbf{Model}}  & \multicolumn{2}{c}{\textbf{HotPot}} & \multicolumn{2}{c}{\textbf{2Wiki}}  & \multicolumn{2}{c}{\textbf{Pop}} \\
    \cmidrule(lr){2-3} \cmidrule(lr){4-5}  \cmidrule(lr){6-7} 
     & EM & F1 & EM & F1 & EM & F1 \\
    \midrule
    Llama3-8b & 0.216 & 0.296 & 0.248 & 0.295 & 0.233 & 0.276 \\
    Qwen2.5-7b & 0.182 &  0.249 & 0.246 & 0.287 & 0.159 & 0.195 \\
    GLM4-9b & 0.190 & 0.262 &  0.245 & 0.308 & 0.185 & 0.235  \\
    Qwen2.5-32b & 0.233 & 0.331 & 0.243 & 0.317 & 0.190 & 0.237 \\
    GPT4o-mini & 0.315 & 0.426 & 0.383 & 0.443 & 0.355 & 0.401\\
    \midrule
    \textbf{Know³-RAG} & \multicolumn{6}{l}{} \\
    \quad GLM4-9b & 0.235 & 0.351 & 0.261  & 0.362  & 0.337  & 0.399 \\
    \quad Qwen2.5-32b & 0.270   &  0.387  & 0.343  & 0.424  & 0.332  & 0.406 \\
    \quad GPT4o-mini & 0.335  & 0.453  & 0.382  & 0.457  & 0.396  & 0.462 \\
    \bottomrule
  \end{tabular}
  }
  \vspace{-1em}
\end{wraptable}

\begin{figure}[t]
    \vspace{-1em}
    \centering
    \includegraphics[width=\textwidth]{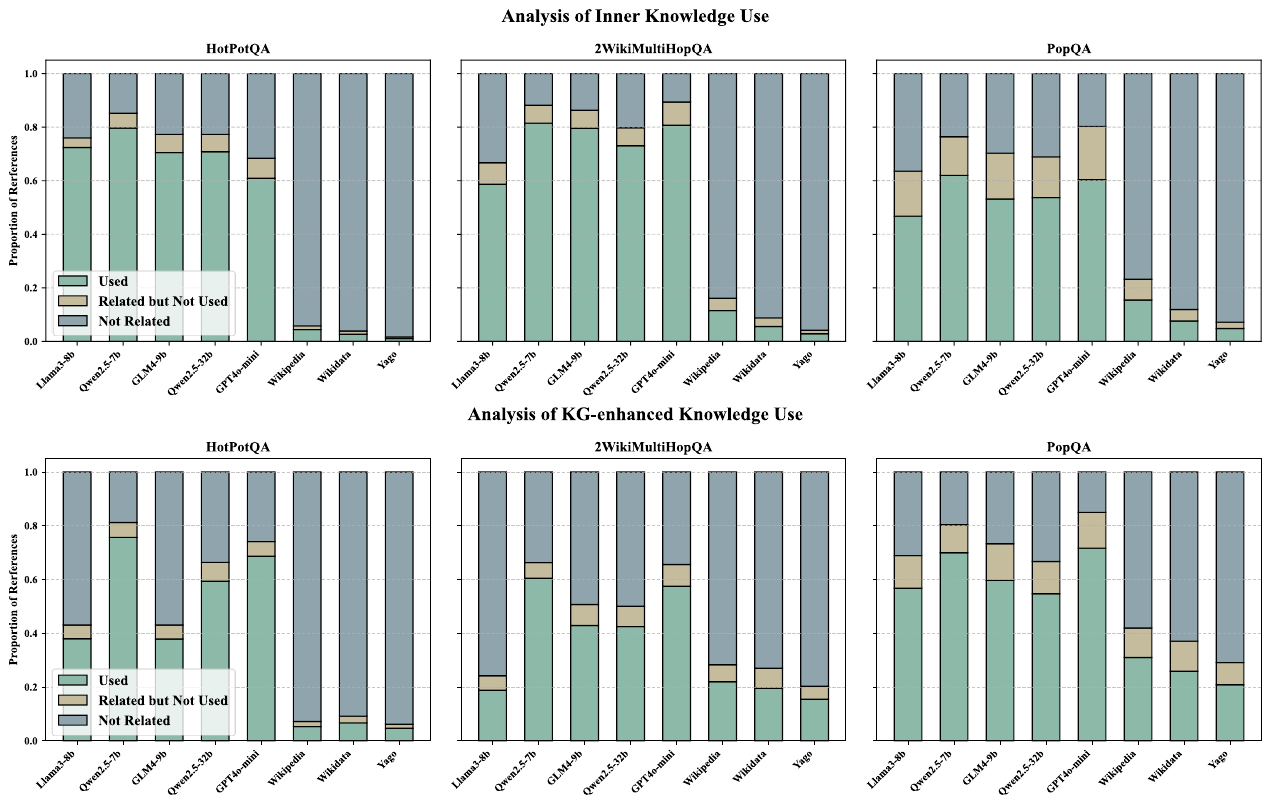}
    \caption{Reference Used of Various Knowledge Models in Turn 1}
    \label{fig:knowledge use}
    \vspace{-1.5em}
\end{figure}

To demonstrate how Know³-RAG utilizes knowledge across models, we first analyze the relevance and usage of references generated by each model. Relevance denotes whether a reference is semantically aligned with the question, while usage indicates whether it is included in the final reference set. As shown in Figure~\ref{fig:knowledge use}, larger models (e.g., LLaMA3, Qwen2.5) exhibit high relevance and usage, reflecting their strong inherent knowledge, consistent with previous ablation results. KG-enhanced queries on these models also yield substantial usage rates. In contrast, smaller models yield low relevance and usage when queried directly, as also observed in the main results of Knowledge Card (KC). However, incorporating KGs significantly boosts their usage, underscoring the role of KGs in stimulating the knowledge of weaker models.
To validate that Know³-RAG selectively exploits correct model knowledge rather than simply aggregating it, Table~\ref{tab:knowledge base} reports QA performance of partial knowledge models. Smaller models are omitted due to fine-tuning limitations. As shown, all models perform worse than Know³-RAG, suggesting effective knowledge selection of Know³-RAG. The impact of the maximum reference is further analyzed in Appendix~\ref{app:ref num}.

\subsection{Analysis of Adaptive Retrieval Control}
To intuitively show how our knowledge-aware adaptive retrieval controls the output decision, Table~\ref{tab:adaptive} presents the percentage of output answers for each turn. 
We observe that only a small percentage of questions are resolved without retrieval at $t = 0$, highlighting the necessity of incorporating external knowledge. At $t = 1$, the majority of questions from HotpotQA and 2WikiMultiHopQA are successfully answered, suggesting that a single retrieval iteration is typically sufficient for these datasets as we fuse the related entities. In contrast, a considerable number of PopQA questions continue to the next turn. We attribute this to the long-tailed nature of PopQA, which makes it challenging for the KGE model to assess the reliability of the responses. 
We also explore the impact of the maximum number of iterations on performance in Appendix~\ref{app:performance with turn}. 

\begin{table}[t] 
  \centering
  \caption{Percentage of Output Answers in Different Turns} 
  \label{tab:adaptive} 
  \setlength{\tabcolsep}{7.5pt} 
    {\small
  \begin{tabular}{@{}l |c| c| c| c| c| c| c| c| c@{}} 
    \toprule
     \multirow{2}{*}[-0.7ex]{\textbf{Turn}} & \multicolumn{3}{c}{\textbf{GLM4-9b}} & \multicolumn{3}{c}{\textbf{Qwen2.5-32b}} & \multicolumn{3}{c}{\textbf{GPT4o-mini}} \\
    \cmidrule(lr){2-4} \cmidrule(lr){5-7} \cmidrule(lr){8-10} 
      & Hotpot & 2Wiki & Pop & Hotpot & 2Wiki & Pop & Hotpot & 2Wiki & Pop \\
    \midrule
    t = 0  &  0.0952 & 0.0166  & 0.0001 & 0.0042 & 0.0009 &  0.0000 &  0.3554 & 0.0180 &  0.0001 \\
    t = 1  &  0.3011 & 0.8384 &  0.0390 &  0.7494 & 0.2253 & 0.0036 & 0.5286 & 0.8611 & 0.0020 \\
    t = 2  &  0.6037 &  0.1450 &  0.9609 &  0.2464 & 0.7738 & 0.9964 &  0.1160 &  0.1209 &  0.9979 \\
    
    \bottomrule
  \end{tabular}
  }
\end{table}



\subsection{Case Study}
\begin{figure}[t]
    \centering
    \includegraphics[width=\textwidth]{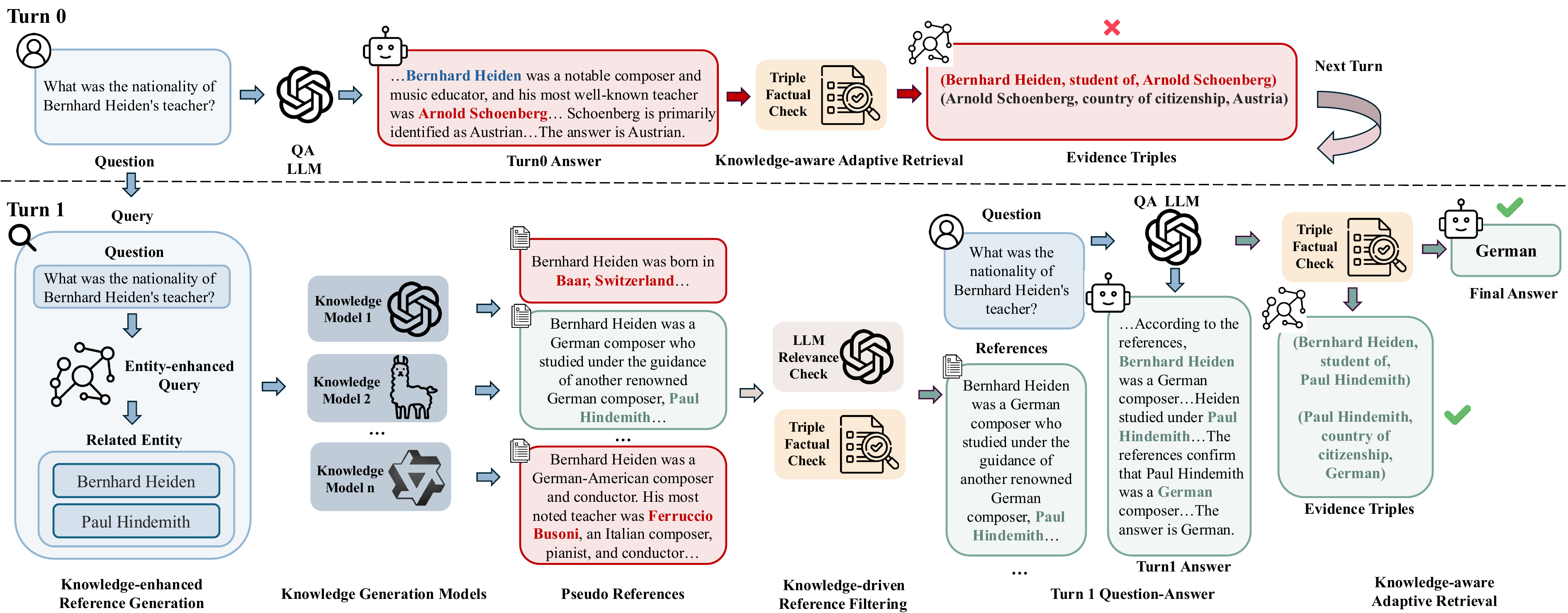}
    \caption{A case study of Know³-RAG, which utilize knowledge graph for open-domain question.}
    \label{fig:case_study}
    \vspace{-0.3cm}
\end{figure}

To provide an intuitive illustration of our framework, we present a representative case in Figure~\ref{fig:case_study}.
At turn 0, the QA model initially produces an incorrect answer, incorrectly assuming that \textit{Bernhard Hayden} studied under \textit{Arnold Schoenberg}. Our Knowledge-aware Adaptive Retrieval catches the error triple \textit{(Bernhard Heiden, student of, Arnold Schoenberg)}, thereby triggering the retrieval.
In turn 1, our framework uses the Knowledge-enhanced Reference Generation module to construct a KG-enhanced query. Specifically, \textit{Bernhard Hayden} and \textit{Paul Hindemith} are identified as related entities. After generating multiple candidate references using the query, our Knowledge-driven Reference Filter filters the irrelevant or incorrect references.
With the filtered references, the QA model re-generates an answer, correctly identifying that \textit{Bernhard Heiden} studied under \textit{Paul Hindemith}, a \textit{German} national. The corresponding factual triples \textit{(Bernhard Heiden, student of, Paul Hindemith)} and \textit{(Paul Hindemith, country of citizenship, German)} are successfully validated, and the correct answer is accepted by our framework.

\section{Conclusions}
In this work, we present Know³-RAG, a knowledge-aware retrieval-augmented generation framework that introduces structured knowledge supervision from knowledge graphs across three key stages: adaptive retrieval, reference generation, and reference filtering. By incorporating external knowledge signals into each component, Know³-RAG enables more reliable control over retrieval behavior, improves the relevance and coverage of retrieved content, and enhances the factual consistency of final outputs. Extensive experiments on multiple open-domain QA benchmarks demonstrate that Know³-RAG consistently outperforms strong baselines in terms of both accuracy and hallucination reduction. We hope our work will lead to more future studies.

\newpage

\bibliography{ref}
\bibliographystyle{ACM-Reference-Format}

\newpage
\appendix

\section{Related Entity Search Algorithm}
\label{appendix:algorithm}
In section~\ref{method:reference generation}, we propose to augment the query with relevant entities from the knowledge graphs. Due to space limitations, the detailed algorithm is presented in the Appendix. As illustrated in Algorithm~\ref{alg:algorithm1}, the process begins by identifying entities \( E_t^q \) within the query via entity linking. Subsequently, based on entities \( E_t^q \), we retrieve related entities from the knowledge graph. Specifically, we first filter relations related to the query based on semantic similarity. Based on these relevant relations, we obtain the local neighbor entities \( E_t^l \) through direct KG querying and global predicted entities \( E_t^g \) via KGE model prediction. Finally, these retrieved entities are incorporated into the query to enrich its context.
\begin{algorithm}[h]
\caption{Related Entity Search.}
	\label{alg:algorithm1}
	\KwIn{Query $q_t$.}
	\KwOut{Related entities $E_t$}  
	\BlankLine
	
    $E^l_t, E^g_t \leftarrow \emptyset, \emptyset$
    \BlankLine
    \tcp{Get the entities in the query through entity linking}
    $E^q_t \leftarrow EL(q_t)$
    \BlankLine
    \tcp{Get the relevant neighboring entities}
    \For{$e^q_t$ in $E^q_t$}{
        \tcp{Query relations around $e^q_t$ in KG}
        $R^{e^q_t}\leftarrow KG(e^q_t)$
        \BlankLine
        \tcp{Obtain the most related tail entity from KG.}
        $r^{e^q_t} \leftarrow max_{r \in R^{e^q_t}}(cos([e^q_t;r], q_t))$
        
        $e^l_t \leftarrow KG(e^q_t, r^{e^q_t})$
        
        $E^l_t \leftarrow E^l_t \cup \{e^l_t\}$
        \BlankLine
        \tcp{Obtain the most related tail entity based on KG embedding.}
        $R^{e^q_t} \leftarrow Topk_{r \in R^{e^q_t}} (cos([e^q_t;r], q_t))$

        $e^g_t \leftarrow max_{r \in R^{e^q_t}}(KGE(e^q_t, r))$

        $E^g_t \leftarrow E^g_t \cup \{e^g_t\}$
    }

    $E_t \leftarrow E^q_t \cup E^l_t \cup E^g_t $

\end{algorithm}

\section{Experimental Supplement}
\subsection{Hyperparameter settings}
\label{app:theta}

As discussed in Section~\ref{method:adaptive}, our adaptive retrieval employs an iterative process with a dynamically increasing threshold \( \theta_t \) to control answer output. \( \theta_t \) is determined by \( \theta_0 \) and a constant c. Due to space limitations, we report the hyperparameter \( \theta_0 \) in Table~\ref{table:theta} in the Appendix.

\begin{table}[thbp] 
    \centering 
    \caption{\( \theta\) for Different Models Across Datasets} 
    \label{table:theta} 
    \begin{tabular}{lccc}
        \toprule
        \textbf{Dataset} & \textbf{GLM4-9b} & \textbf{Qwen2.5-32b} & \textbf{GPT4o-mini}\\
        \midrule
        HotpotQA & 10 & 1 & 13\\
        2WikiMultiHopQA & 2 & 0.2 & 2 \\
        PopQA & 0.1 & 0.02 & 0.01\\
        \bottomrule
    \end{tabular}
\end{table}



\begin{table}[]
    \centering
  \caption{Performance with Varied References} 
  \label{tab:knowledge num} 
  \begin{tabular}{@{}c |cc| cc| cc@{}} 
    \toprule
    & \multicolumn{6}{c}{GLM4-9b} \\
     \multicolumn{1}{@{}l|}{Method}  & \multicolumn{2}{c}{HotPot} & \multicolumn{2}{c}{2Wiki}  & \multicolumn{2}{c}{Pop} \\
    \cmidrule(lr){2-3} \cmidrule(lr){4-5}  \cmidrule(lr){6-7} 
     & EM & F1 & EM & F1 & EM & F1 \\
    \midrule
    1 & 0.196 & 0.296 & 0.256 & 0.329 & 0.320 & 0.376 \\
    3 & 0.215 & 0.328 & 0.231 & 0.320 &  0.325 & 0.389 \\
    5 & 0.227 & 0.340 & 0.260 & 0.358 & 0.331 & 0.391 \\
    7 & 0.217 & 0.333 & 0.216 & 0.309 & 0.333 &  0.396 \\
    9 & 0.225 & 0.337 & 0.217 & 0.311 & 0.332 &  0.396 \\
    \bottomrule
  \end{tabular}
\end{table}

\subsection{Impact of maximum reference number}
\label{app:ref num}
To assess the influence of the maximum number of references on model performance, we vary the number of documents provided to the QA model during the first turn and report results across three datasets in Table~\ref{tab:knowledge num}. Performance generally improves as the number of references increases, but the gains reduce or even decline beyond a certain point. This suggests that although reference filtering is applied, introducing too many documents can still lead to irrelevant or redundant content, ultimately hindering answer quality.

\subsection{Impact of the Maximum Number of Iterations}
\label{app:performance with turn}

\begin{figure}[t]
    \centering
    \includegraphics[width=\textwidth]{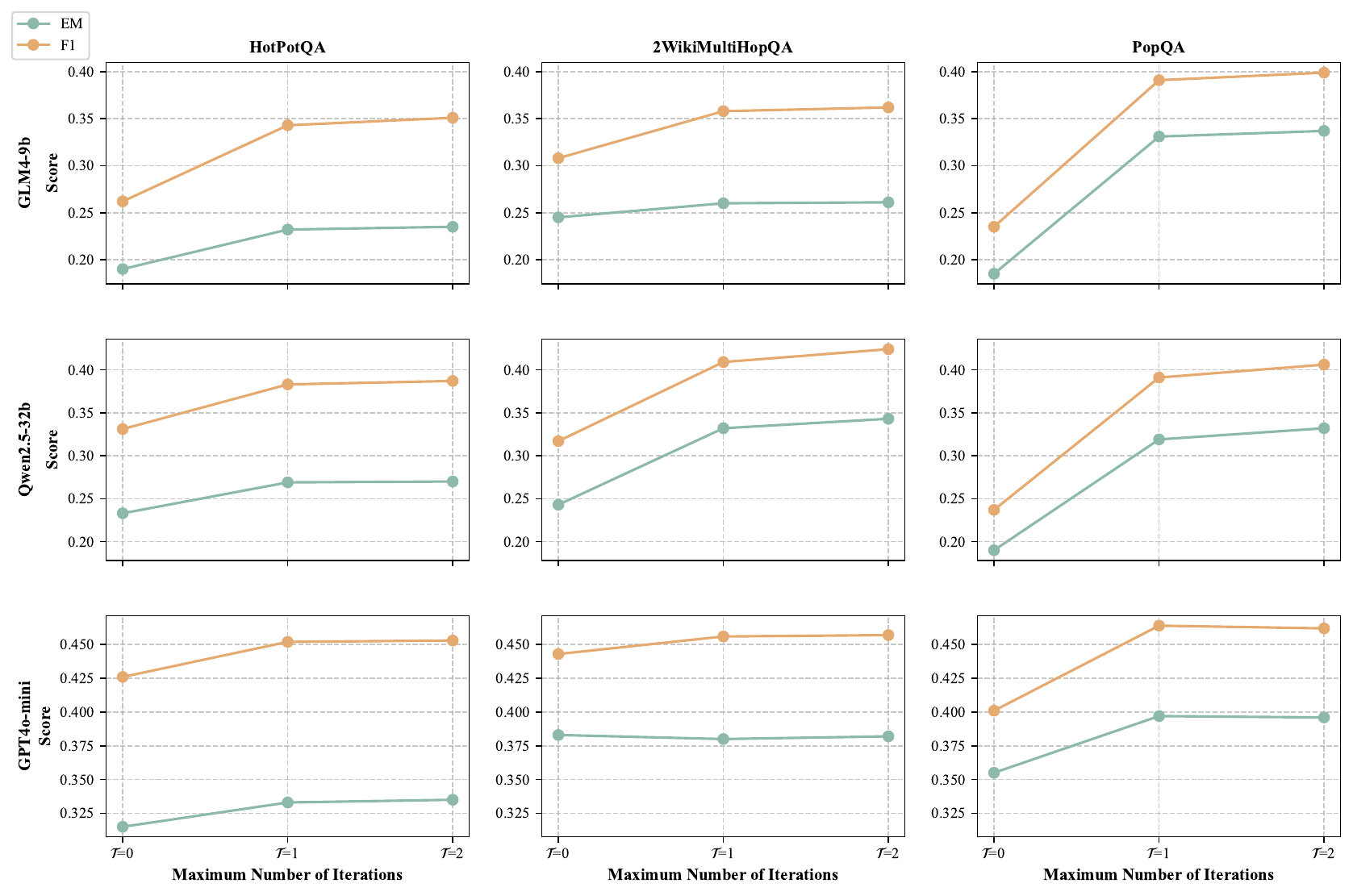}
    \caption{Performance of Know³-RAG with Varying Maximum Number of Iterations}
    \label{fig:reference num}
\end{figure}

To explore the impact of the maximum number of turns on model performance, Figure~\ref{fig:reference num} shows the performance with the varying maximum number of iterations. We notice that all models demonstrate higher EM and F1 scores at $\mathcal{T} = 2$ compared to $\mathcal{T} = 0$, indicating that iterative retrieval positively impacts answer quality.
We also find that the growth from $\mathcal{T} = 0$ to $\mathcal{T} = 1$ is significantly greater than from $\mathcal{T} = 1$ to $\mathcal{T} = 2$, indicating that most questions can be effectively addressed after the first retrieval turn. This finding is consistent with Table~\ref{tab:adaptive} and further demonstrates the effectiveness of our knowledge-aware adaptive retrieval.

\section{Limitation}
\label{limiation}
 While our framework demonstrates promising results, several limitations deserve discussion.
First, the integration of knowledge graphs introduces non-negligible noise. For efficiency, we adopt relatively simple methods for triple extraction and entity linking, which may introduce noise into the retrieved entity knowledge. This noise, in turn, reduces the relative contribution of external knowledge in the overall performance. Future work could incorporate more accurate extraction and linking techniques to enhance knowledge quality~\cite{wu2019blink,ayoola2022refined}.
Second, the source of the retrieved text can be further expanded. Our current framework primarily relies on knowledge graphs due to their high reliability. However, knowledge graphs are often updated with some delay, which may hinder the timeliness of retrieved facts. To address this, incorporating search engine-based retrieval as a supplementary source presents a viable direction~\cite{liu2023webglm,feng2024knowledge-card}. Nevertheless, such sources often include irrelevant information and require advanced denoising, which we leave for future exploration.

\section{Prompts}
\label{Appendix:prompt}
In this section, we show the prompt we used for the question answer, reference generation, query generation, triple extraction, and relevance check.

\begin{figure}[th]
    \centering
    \includegraphics[width=\textwidth]{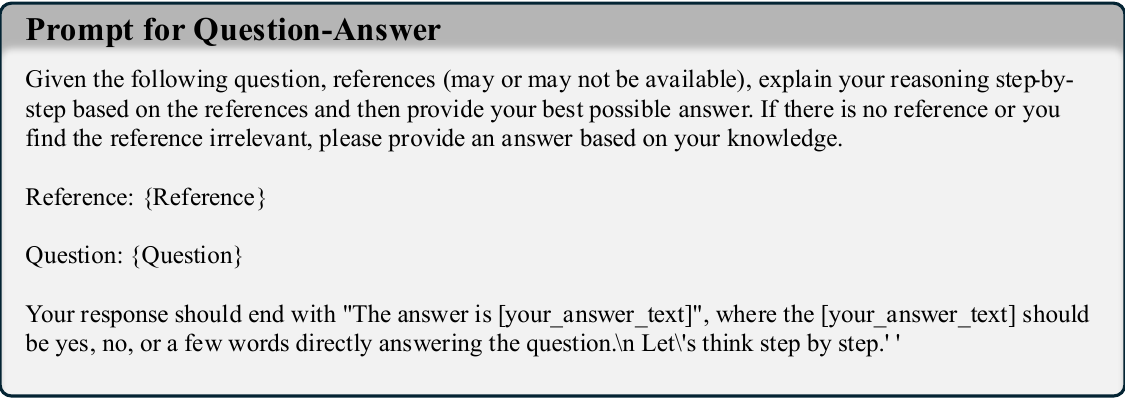}
    \caption{Prompt for Question Answer}
    \label{fig:rel prompt}
\end{figure}
\begin{figure}[th]
    \centering
    \includegraphics[width=\textwidth]{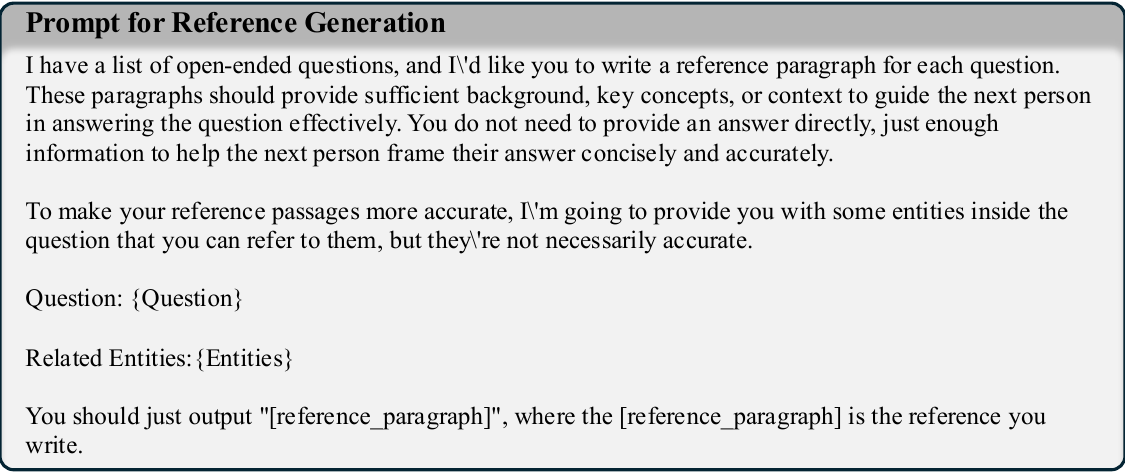}
    \caption{Prompt for Reference Generation}
    \label{fig:rel prompt}
\end{figure}
\begin{figure}[th]
    \centering
    \includegraphics[width=\textwidth]{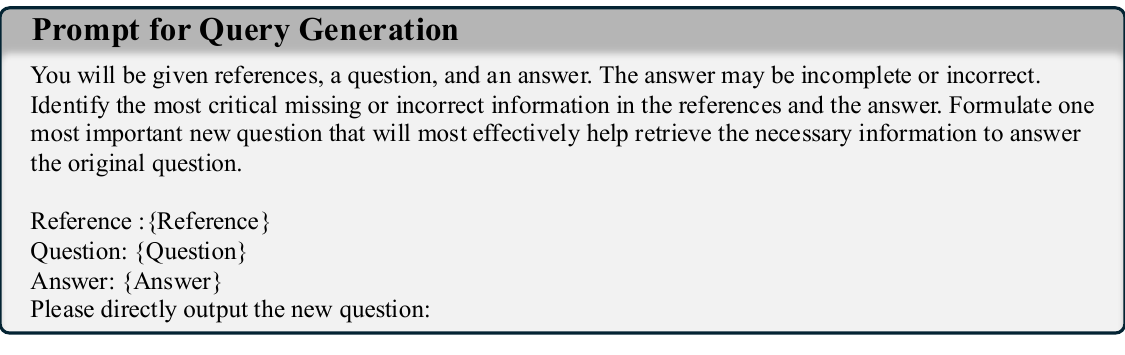}
    \caption{Prompt for Query Generation}
    \label{fig:rel prompt}
\end{figure}
\begin{figure}[th]
    \centering
    \includegraphics[width=\textwidth]{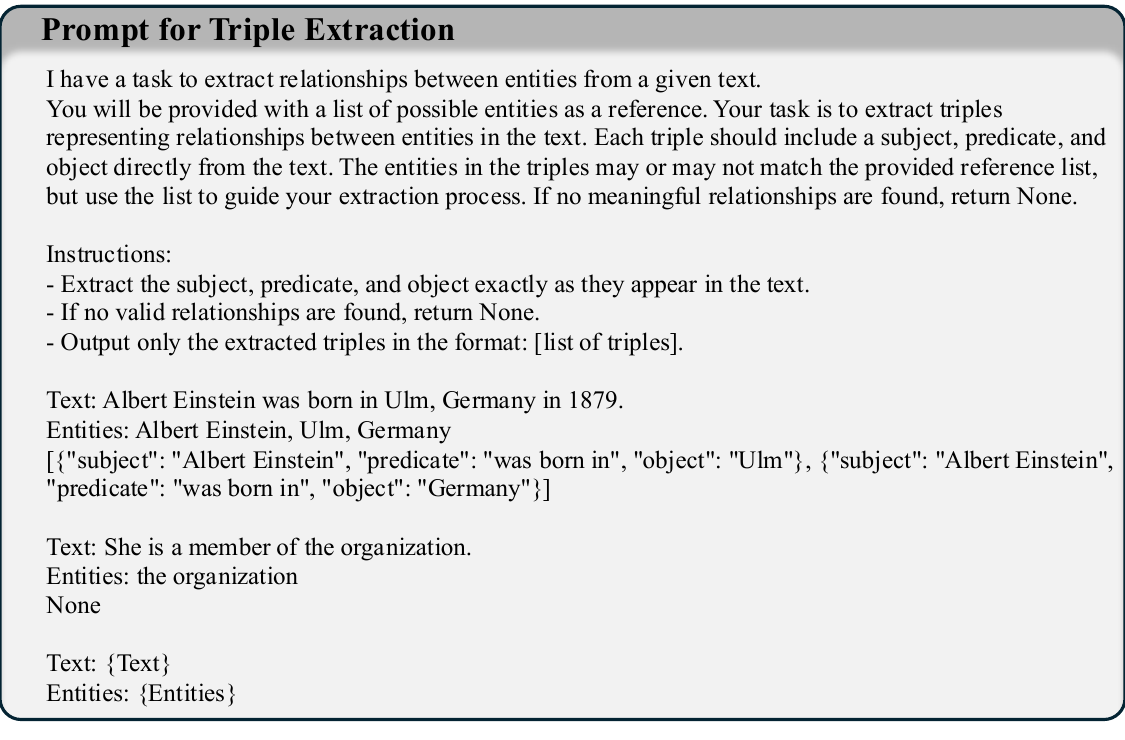}
    \caption{Prompt for Triple Extraction}
    \label{fig:rel prompt}
\end{figure}
\begin{figure}[th]
    \centering
    \includegraphics[width=\textwidth]{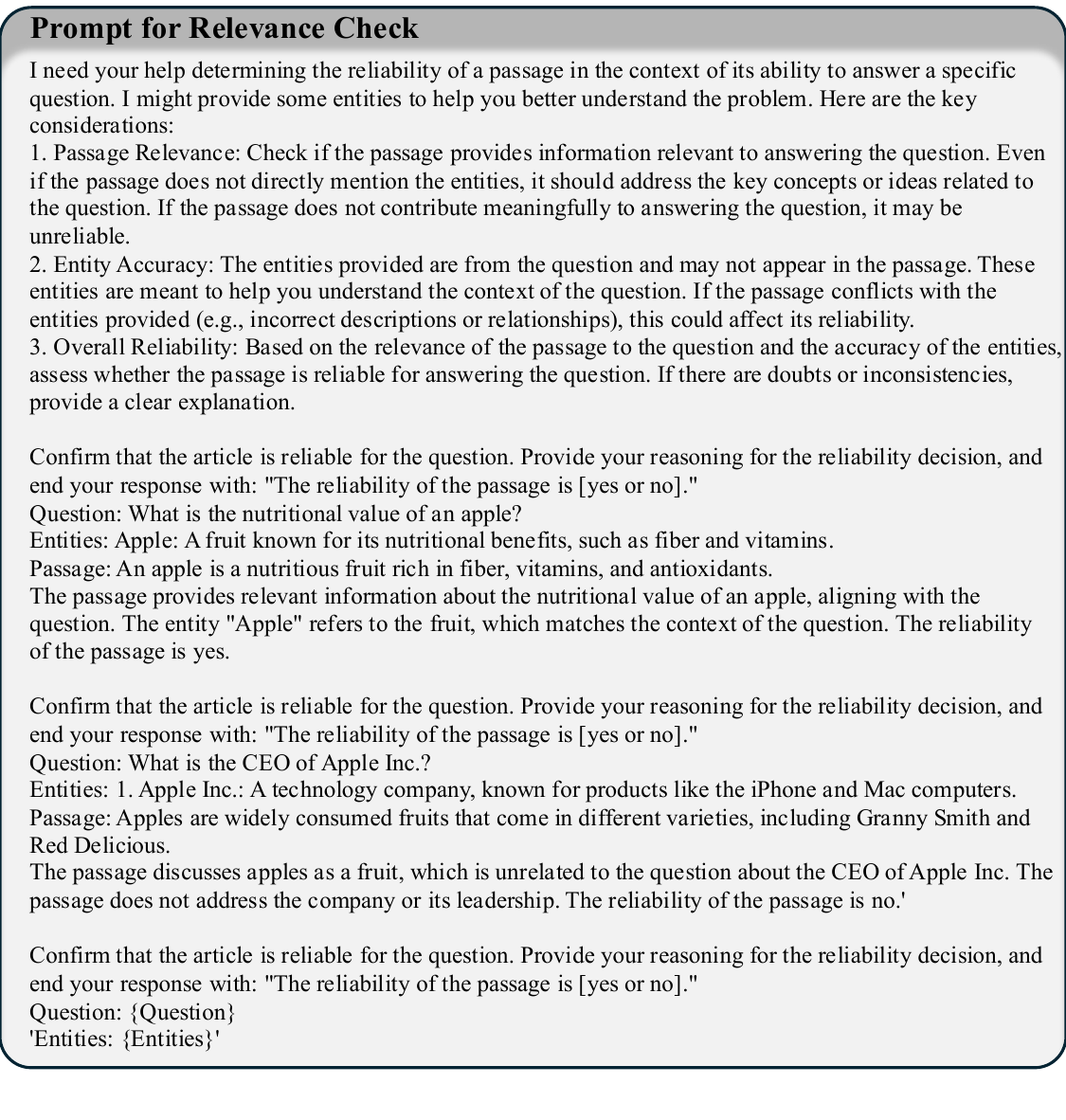}
    \caption{Prompt for Relevance Check}
    \label{fig:rel prompt}
\end{figure}

\end{document}